\documentclass[10pt,twocolumn,letterpaper]{article}
\usepackage{cvpr}              
\usepackage[accsupp]{axessibility} 
\usepackage{algorithm,algorithmic}
\usepackage{multirow}
\usepackage{amsmath}
\newtheorem{definition}{\textbf{\emph{Definition}}}

\usepackage[dvipsnames]{xcolor}

\definecolor{cvprblue}{rgb}{0.21,0.49,0.74}
\usepackage[pagebackref,breaklinks,colorlinks,citecolor=cvprblue]{hyperref}

\def\paperID{11003} 
\def\confName{CVPR}
\def\confYear{2024}

\title{S$^2$MVTC: a Simple yet Efficient Scalable Multi-View Tensor Clustering}

\author{
Zhen Long\thanks{Equal contribution.}
, Qiyuan Wang\footnotemark[1]
, Yazhou Ren
, Yipeng Liu
, Ce Zhu\thanks{Corresponding author.This work is supported by the National Natural Science Foundation of China (NSFC) under Grant 62020106011.}
\\
University of Electronic Science \& Technology of China\\
{\tt\small eczhu@uestc.edu.cn}
}

\begin{document}
\maketitle

\begin{abstract}
Anchor-based large-scale multi-view clustering has attracted considerable attention for its effectiveness in handling massive datasets. 
However, current methods mainly seek the consensus embedding feature for clustering by exploring global correlations between anchor graphs or projection matrices.
In this paper, we propose a simple yet efficient scalable multi-view tensor clustering (S$^2$MVTC) approach, where our focus is on learning correlations of embedding features within and across views. 
Specifically, we first construct the embedding feature tensor by stacking the embedding features of different views into a tensor and rotating it. Additionally, we build a novel tensor low-frequency approximation (TLFA) operator, which incorporates graph similarity into embedding feature learning, efficiently achieving  smooth representation of embedding features within different views. 
Furthermore, consensus constraints are applied to embedding features to ensure inter-view semantic consistency. 
Experimental results on six large-scale multi-view datasets demonstrate that S$^2$MVTC significantly outperforms state-of-the-art algorithms in terms of clustering performance and CPU execution time, especially when handling massive data. 
The code of S$^2$MVTC is publicly available at https://github.com/longzhen520/S2MVTC.
\end{abstract}    
\section{Introduction}
\label{sec:intro}



The progress in information collection technology now permits us to gather multi-view data from the same object, presenting observations with depth and comprehensiveness.
For instance, brain activity can be measured using functional Magnetic Resonance Imaging (fMRI) and Electroencephalography (EEG)~\cite{huster2012methods,mulert2023eeg}.
Benefiting from the consensual and complementary information, multi-view data have attracted a series of multi-view learning tasks~\cite{sun2013survey,baltruvsaitis2018multimodal}. Among them, multi-view clustering, which groups data into several clusters by integrating information from different views, has been extensively applied in fields including image processing, computer vision, and neuroscience~\cite{serra2024multiview,wang2022multi,busch2023multi,yang2022robust}.

Current multi-view clustering (MVC) methods, mainly based on self-representation learning or graph learning, aim to find the consensus embedding feature and
have attained considerable advancements~\cite{chen2022low,si2022consistent,li2023auto,10225718}. 
However, for $V$ views and $N$ samples, these methods require updating $V$ membership graphs ($N \times N$) to construct the affinity matrix, which will be then fed into spectral clustering algorithm~\cite{von2007tutorial}. 
Both storage and computational demands for these methods scale at a complexity of $O(VN^3)$, where $V$ and $N$ represent the numbers of views and samples, respectively. These methods could be infeasible for large-scale datasets, particularly when $N$ is massive, and such scalability is important in real-world applications.

\begin{figure}[htbp]
 \vspace{-0.31cm}
 \centering
\includegraphics[width=0.485\textwidth]{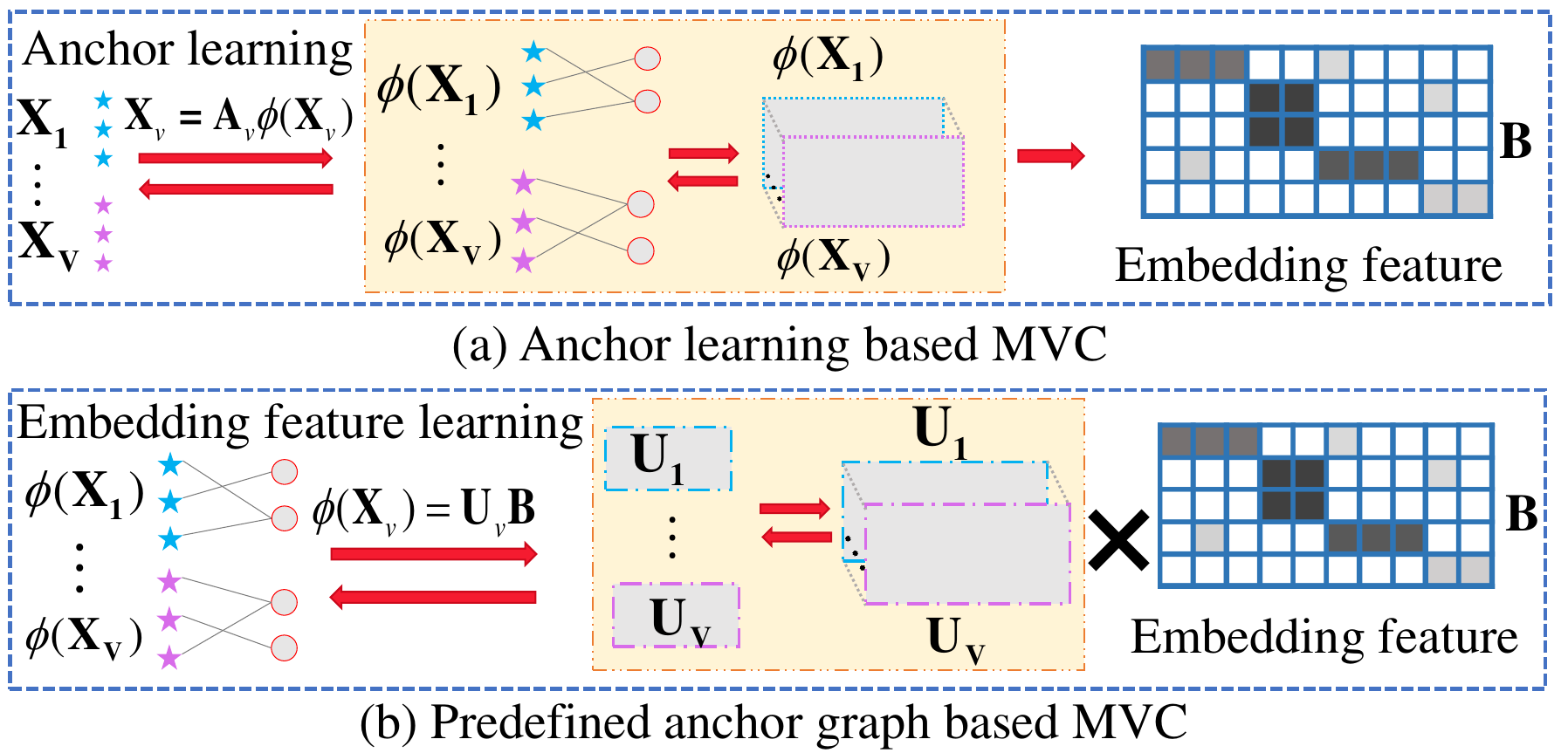}
 \vspace{-0.8cm}
\caption{Comparison frameworks of current methods.}
\label{fig:rebuttal2}
\vspace{-0.2cm}
 \end{figure}

To address it, many anchor-based scalable MVC methods have been proposed for large-scale data~\cite{xia2022tensorized,li2020multiview,shu2022self,wang2021fast}, where $V$ anchor graphs of size $M\times N$ are constructed from $M$ anchors and $N$ samples ($M\ll N$) to approximately represent $V$ memberships between multi-view data. 
According to the differences in inter-view processing levels, these methods can be further subdivided into two categories. 
The first one relies on anchor learning, emphasizing the exploration of anchor graph consistency across views, as shown in \cref{fig:rebuttal2} (a). Subsequently, the fused anchor graph is employed to construct the embedding feature, which will be fed into a $k$-means algorithm for clustering~\cite{9646486,xia2022tensorized,Ji_2023_ICCV,10225718}. 
The second one relies on provided anchor graphs and learns the embedding feature using the projection matrices, as shown in \cref{fig:rebuttal2} (b).
It explores the consistency among projection matrices to acquire the compact embedding feature, which is then utilized for learning clustering structures~\cite{9882008,10032273,zhang2018binary}.
\begin{figure*}
    \centering
    \includegraphics[width=1\linewidth]{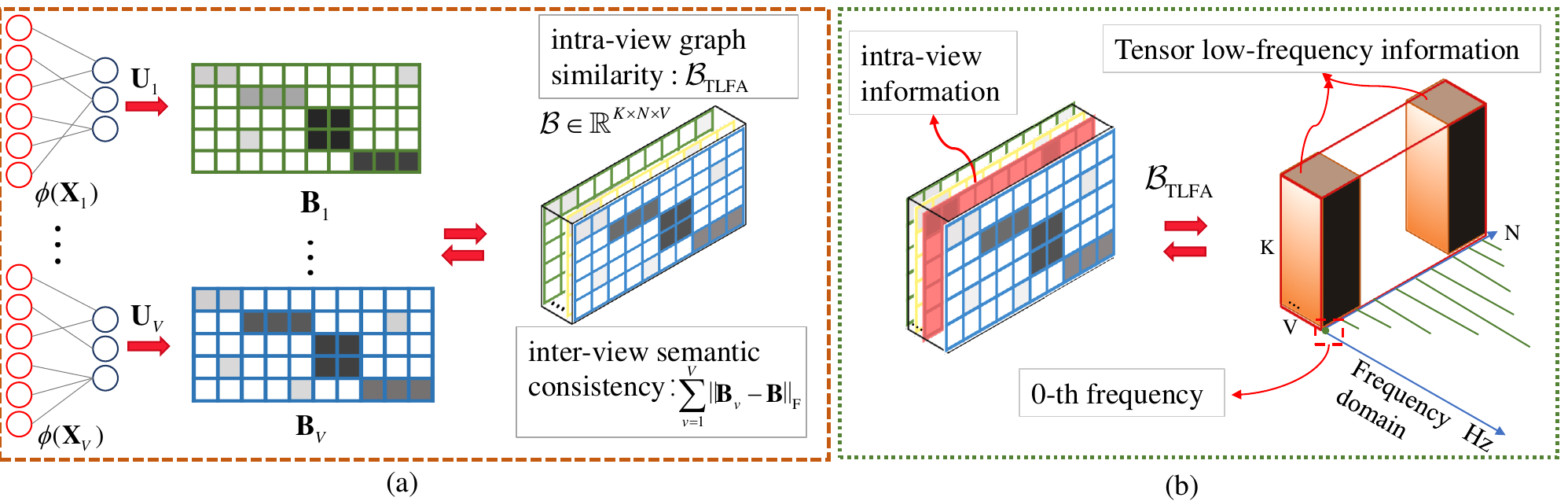}
    \caption{The framework of S$^2$MVTC. 
    (a) The process involves mapping each pre-given anchor graph $\phi(\mathbf{X}_v)$ using $\mathbf{U}_v$ to obtain the corresponding embedding feature $\mathbf{B}_v$. These embedding features, $\mathbf{B}_v$, are then formed into a tensor $\mathcal{B}$. This tensor, along with the newly defined $\mathcal{B}_\textbf{TLFA}$ and $\sum_{v=1}^{V}\|\mathbf{B}_v-\tilde{\mathbf{B}}\|_{\operatorname{F}}$, are utilized to explore intra-view  correlations and  semantic consistency, respectively, where $\tilde{\mathbf{B}}=\sum_{v=1}^{V}\frac{1}{V}\mathbf{B}_v$. 
    (b) The tensor low-frequency approximation $\mathcal{B}_\textbf{TLFA}$.
    }
    \label{fig:framework}
\end{figure*}

The above-mentioned anchor-based methods either explore the global correlations between anchor graphs or projection matrices. However, their ultimate goal is to learn the consensus embedding feature, which directly impacts the clustering performance.
Therefore, two questions naturally arise: Why not directly explore the correlations between embedding features from different views? Would this approach be more effective?

In this paper, we try to answer these questions and focus on learning inter-and intra-view consistency among embedding features.
We propose a simple yet efficient scalable multi-view tensor clustering (S$^2$MVTC) approach tailored for large-scale data. 
 As depicted in  \cref{fig:framework}, we initially map the anchor graph $\phi(\mathbf{X}_v)\in\mathbb{R}^{M\times N}$ onto the projection matrix $\mathbf{U}_v$ to obtain the embedding feature $\mathbf{B}_v\in\mathbb{R}^{K\times N}$ for each view. 
 To ensure the consensus among intra-view embedding features effectively, we introduce a novel approach by employing a tensor low-frequency approximation (TLFA) on the embedding feature tensor, denoted as $\mathcal{B}\in\mathbb{R}^{K\times V\times N}$, where the tensor is formed by concatenating the embedding features from each view and subsequently rotating it.
 Benefiting from the fast Fourier transform (FFT) in the third mode of tensor singular value decomposition (t-SVD)~\cite{kilmer2013third}, the newly defined TLFA can achieves the smooth representation of samples within different views~\cite{hu2014smooth}.  

Furthermore, to ensure inter-view semantic consistency, a consensus constraint is imposed on the embedding features. By incorporating these two parts into a unified framework, S$^2$MVTC can efficiently leverage inter/intra-view information for large-scale MVC tasks. Ultimately, the clustering structure learning is applied to the well-learned embedding consensus feature $\Tilde{\mathbf{B}}$ to obtain the clustering results.\\
The main contributions are summarized as:
 \begin{itemize}
    \item  Different from existing anchor-based methods that explore global correlations between anchor graphs or projection matrices, S$^2$MVTC directly learns both inter and intra-view embedding feature correlations.
    \item  Benefiting from the newly defined TLFA operator, S$^2$MVTC achieves the smooth representation within different views.
     \item  
     Experimental results on six large multi-view datasets demonstrate that S$^2$MVTC significantly improves the clustering performance compared to state-of-the-art algorithms, especially as the data size increases, making the advantages of S$^2$MVTC more evident.
 \end{itemize}



\section{Notations and Problem Formulation}
\subsection{Notations}
For clarity, we present notations frequently used in \cref{tab:notatons}.
\begin{table}[htbp]
\centering
\caption{Summary of notations in this paper.}
\scalebox{0.8}{
    \begin{tabular}{{cc}}
        \toprule
Symbol   & Definition \\
        \midrule
        $x$, $ \mathbf{x} $,  $  \mathbf{X} $, and $\mathcal{X}$& A scalar, a vector, a matrix, and a tensor\\
        $v = 1,\cdots, V$& Indices  range from 1 to their capital version\\
        $\mathbb{R},\mathbb{C}$ & Fields of real numbers and complex numbers\\
        $\operatorname{conj}(\mathcal{X})$ & the complex conjugate of $\mathcal{X}$\\
        $\lceil x\rceil$ & the one greater than or equal to $x$\\
        $\mathcal{X}\in \mathbb{R}^{I_1\times I_2 \times I_3}$& A 3-rd order tensor\\
        $\mathcal{X}(:,:,i_3)$& the $i_3$-th frontal slice of $\mathcal{X}$\\
$N,V,C,M$                                                & Number of samples, views, clusters, anchors\\
$\phi(\mathbf{X}_v)\in \mathbb{R}^{M\times N}$             &   Anchor graph in the $v$-th view  \\ 
$\mathbf{U}_v\in\mathbb{R}^{M\times K}$                                     &   Projection matrix in the $v$-th view\\
$\mathbf{B}_v\in\mathbb{R}^{K\times N}$                                     &   embedding feature in the $v$-th view  \\
$\mathcal{B}\in \mathbb{R}^{K\times V\times N}$    &   The rotated embedding feature tensor \\ 
        \bottomrule 
    \end{tabular}}

\label{tab:notatons}
\end{table}

\subsection{Preliminaries on t-SVD}

\begin{definition}(\textbf{T-product}~\cite{recht2011simpler})\label{definition:t-production}
    The T-product of two third-order tensors $\mathcal{X}\in\mathbb{R}^{I_1\times I_2\times I_3}$ and $\mathcal{Y}\in\mathbb{R}^{I_2\times J\times I_3}$ is defined as 
   \begin{equation}
       \mathcal{Z}=\mathcal{X} * \mathcal{Y}\in\mathbb{R}^{I_1\times J\times I_3},
   \end{equation}
  which can be computed by multiple matrix multiplication operations in the Fourier transform domain.
\end{definition}
\begin{definition}(\textbf{Tensor transpose \cite{kilmer2013third} )}) 
Under the Fourier transform, the conjugate transpose of a tensor $\mathcal{X}\in\mathbb{C}^{I_1 \times I_2 \times I_3}$ is denoted as  $\mathcal{X}^{\operatorname{T}}\in\mathbb{C}^{I_2\times I_1\times I_3}$. It satisfies   
 \begin{equation}
\mathcal{X}^{\operatorname{T}}(:,:,1) =\left(\mathcal{X}(:,:,1)\right)^{\operatorname{T}},
\end{equation}
and 
\begin{equation}
\mathcal{X}^{\operatorname{T}}(:,:,i_3) =\left(\mathcal{X}(:,:,I_3+2-i_3)\right)^\top,i_3=2,...,I_3 ,
   \label{eq:important}
 \end{equation}
 where $\mathcal{X}^{\operatorname{T}}(:,:,i_3)$ is the $i_3$-th frontal slice of $\mathcal{X}^{\operatorname{T}}$.
\end{definition}

\begin{definition}(\textbf{Orthogonal tensor})~\cite{kilmer2013third} \label{def: Orthogonal tensor}
 A tensor $\mathcal{X}$ is called orthogonal when it satisfies
  \begin{equation}
  \mathcal{X}*\mathcal{X}^{\operatorname{T}}=\mathcal{X}^{\operatorname{T}}*\mathcal{X}=\mathcal{I},
  \end{equation}
  where $\mathcal{I}$ is the identity tensor.  
\end{definition}

\begin{definition}(\textbf{f-diagonal tensor})~\cite{kilmer2011factorization} \label{def: F-diagonal tensor}
A tensor $\mathcal{X}\in \mathbb{R}^{I_1 \times I_2 \times I_3} $ is called f-diagonal when  each frontal slice $\mathcal{X}^{(i_3)}, i_3 = 1, \cdots, I_3$ is a diagonal matrix.
\end{definition}  

\begin{definition}(\textbf{tensor Singular Value Decomposition(t-SVD) \cite{kilmer2013third,8606166,braman2010third})})
For a tensor $\mathcal{X} \in \mathbb{R}^{I_1 \times I_2 \times I_3}$, its t-SVD is represented as:
 \begin{equation}
\mathcal{X} = \mathcal{U} * \mathcal{S} * \mathcal{V}^{\operatorname{T}},
   \label{eq:important}
 \end{equation}
where $\mathcal{U}\in\mathbb{R}^{I_1\times I_1\times I_3}$ and $\mathcal{V}\in\mathbb{R}^{I_2\times I_2\times I_3}$ are orthogonal tensors, and $\mathcal{S}\in\mathbb{R}^{I_1\times I_2\times I_3}$ is an f-diagonal tensor. 
\end{definition}
According to the T-product operation in  \cref{definition:t-production}, this decomposition can be obtained by  \cref{alg:tsvd}, where $\mathrm{fft}$ and $\mathrm{ifft}$ are the fast Fourier transform (FFT) operator and the inverse FFT, respectively.

\begin{algorithm}[t]
	\caption{t-SVD}	
           	\begin{algorithmic}
		\STATE \textbf{Input:} $\mathcal{X} \in \mathbb{R}^{I_1 \times I_2 \times I_3}$.
               \STATE \textbf{Output:} $\mathcal{U} \in \mathbb{R}^{I_1 \times I_1 \times I_3}$, $\mathcal{S} \in \mathbb{R}^{I_1 \times I_2 \times I_3}$, $\mathcal{V} \in \mathbb{R}^{I_2 \times I_2 \times I_3}$;
			 \STATE $\widehat{\mathcal{X}}\leftarrow\mathrm{fft}(\mathcal{X},[],3)$;
            \STATE $[\mathbf{U}, \mathbf{S}, \mathbf{V}] = \mathbf{svd}(\widehat{\mathcal{X}}(:,:,1))$.
            \FOR{$i_3 = 1$ to $ I_3$}
            \STATE $[\mathbf{U}, \mathbf{S}, \mathbf{V}] = \mathbf{svd}(\widehat{\mathcal{X}}(:,:,i_3))$;
            \STATE $\widehat{\mathcal{U}}(:,:,i_3) =\mathbf{U}; \widehat{\mathcal{S}}(:,:,i_3) =\mathbf{S}; \widehat{\mathcal{V}}(:,:,i_3)=\mathbf{V}$.
            \ENDFOR
            \STATE $\mathcal{U}=\text{ifft}(\widehat{\mathcal{U}}, [\ ], 3);\mathcal{S}=\text{ifft}(\widehat{\mathcal{S}}, [\ ], 3);\mathcal{V} =\text{ifft}(\widehat{\mathcal{V}}, [\ ], 3).$    \end{algorithmic}	
\label{alg:tsvd}
\end{algorithm}


\subsection{Related works}

\subsubsection{Anchor learning for large-scale MVC }
 The general framework of an anchor-learning-based large-scale MVC is constructed as follows:
 \begin{eqnarray}\label{general model v1}
   &&\min_{\{\mathbf{Z}_v\}_{v=1}^{V}} \sum_{v=1}^{V} \|\mathbf{X}_v-\mathbf{A}_v\mathbf{Z}_v\|_{\operatorname{F}}^2+ \lambda\Psi(\mathbf{Z}_v) \nonumber\\
  &&\text{s.~t.~} \mathbf{Z}_v^{\operatorname{T}}\mathbf{1}=\mathbf{1},\mathbf{Z}_v\geq 0, v=1,\cdots,V,
\end{eqnarray}
where $\mathbf{A}_v\in\mathbb{R}^{D_v\times M}$ is the learned anchor matrix in the $v$-th view and $M$ is the number of anchors. $\mathbf{Z}_v\in\mathbb{R}^{M\times N}$ is the anchor graph, which is used to describe the relationship among samples.  In addition, $\Psi(\mathbf{Z}_v)$ represents the regularization terms.  Kang \etal~\cite{kang2020large} consider  Frobenius norm on $\mathbf{Z}_v$ to find a stable solution for large-scale MVC. However, the anchor graph in each view is learned separately, failing to explore the complementary information. To solve this, many fusion strategies are considered to find the consensus representation among views. For example, Liu \etal~\cite{9646486} 
 considered the respective projection matrix to find the consistent graph. Li \etal~\cite{9146384} considered the joint graph across multiple views
via a self-supervised weighting manner.
Chen \etal~\cite{10087316} introduced the Fast Self-guided Multi-view Subspace Clustering algorithm, which effectively combines view-shared anchor learning and global-guided-local self-guidance learning into a comprehensive model. Huang \etal~\cite{10016684} considered three levels of diversity: features, anchors, and neighbors. By leveraging these levels of diversity,  view-sharing bipartite graphs are constructed, enhancing the effectiveness of the clustering process. In addition, some methods based on tensor anchor graphs have been proposed~\cite{xia2022tensorized,Ji_2023_ICCV,10225718}. These methods stack each anchor graph into a tensor and use low-rank tensor approximation to integrate the consistency information among anchor graphs across views, achieving good performance on large-scale data.

\subsubsection{Embedding feature learning for large-scale MVC}
Unlike large-scale MVC methods based on anchor learning, another approach focuses on learning embedding features through the pre-defined anchor graphs in the pre-processing step. Typically, these anchor graphs are constructed using kernel mapping, for instance, through nonlinear Radial Basis Function (RBF) mapping, as follows:
\begin{equation}
    \phi(\mathbf{x}_v^n)=[\exp(-\frac{\|\mathbf{x}_v^n-\mathbf{z}_v^1\|^2}{\sigma}),\cdots,\exp(-\frac{\|\mathbf{x}_v^n-\mathbf{z}_v^M\|^2}{\sigma})]^{\operatorname{T}},
\end{equation}
where $\sigma$ is the  kernel width, $\mathbf{z}_v^{m}\in\mathbb{R}^{D_v}, m=1,\cdots,M$ are  anchors, which
randomly chooses from the column of $v$-th
view $\mathbf{X}_v\in\mathbb{R}^{D_v\times N}$. $\phi(\mathbf{x}_v^n)\in\mathbb{R}^{M}$ represents the nonlinear relationship between $M$ anchors and the $n$-th samples. 
In this way, Zhang \etal~\cite{zhang2018binary} projected non-linear anchor graphs from different views to a common Hamming feature space. This shared feature space is then fed into binary clustering structure learning to derive the final clustering results. However, this method solely explores pairwise correlations between views.
To uncover global correlations among views,
 Zhang \etal~\cite{9882008} considered consistent projections, obtained by low-rank tensor approximation, to find a better binary feature representation for large-scale MVC tasks. Additionally,  Wang \etal~\cite{10032273} considered autoencoder learning techniques to find consistent projections for generating binary features used in clustering.

\section{Methods}
As mentioned above, anchor-based methods mostly focus on learning high-order information between anchor graphs or projection matrices. However, their ultimate goal is to obtain shared embedding features across views.
\subsection{Model development}
In this section, we will focus on how to quickly learn the consistency among embedding features well, which involves the exploration of intra-view graph similarity  (IGS) and ensuring inter-view semantic consistency (ISC).
With anchor graphs $\phi(\mathbf{X}_v)\in\mathbb{R}^{M\times N}, v=1,\cdots, V$ obtained in advance, we can first learn  $V$ embedding features by projection matrices $\mathbf{U}_v, v=1,\cdots, V$, as follows:
\begin{equation}
\begin{aligned}
&\min_{\{\mathbf{U}_v,\mathbf{B}_v\}_{v=1}^{V}} \sum_{v=1}^{V}\frac{\lambda}{2}||\mathbf{U}_v||_{\operatorname{F}}^2+\frac{1}{2}||\mathbf{B}_v-\mathbf{U}_v\phi(\mathbf{X}_v)||_{\operatorname{F}}^2\\
& \text{s.~t.~} 
\mathbf{B}_v(:,n)\in\mathbb{S}, n=1,\cdots, N, v=1,\cdots,V,
\end{aligned}
\end{equation}
 where $\lambda$ serves as a weight aiming to attain a stable solution for the mapping matrix $\mathbf{U}_v$. $\mathbf{B}_v(:,n)\in\mathbb{S}$ represents the $z$-score normalization constraint, ensuring that each feature from different views has a mean of 0 and a standard deviation of 1. 
Here $\mathbf{b}_v^n=\mathbf{B}_v(:,n)$, and 
$\mathbb{S}=\{\mathbf{b}_v^n:\sum_{k=1}^{K}\mathbf{b}_v^n(k)={0},  \frac{\sum_{k=1}^{K}(\mathbf{b}_v^n(k))^2}{K-1}=1, n=1,\cdots,N\}$.

To efficiently explore the intra-view information of embedding features, we initially stack the embedding features from each view into a tensor. Subsequently, we introduce a newly tensor low-frequency approximation $\|\mathcal{B}\|_{\text{TLFA}}$, applied to its rotated form  $\mathcal{B}\in\mathbb{R}^{K\times V\times N}$, aiming to capture the IGS.
\begin{definition}(\textbf{Tensor low-frequency approximation (TLFA)})
According to the t-SVD computation outlined in  \cref{alg:tsvd}, the TLFA of a tensor $\mathcal{B}\in\mathbb{R}^{K\times V\times N}$ is defined as the frontal
slices in the low-frequency domain. Mathematically, it is expressed as:
\begin{equation*}
\vspace{-0.3cm}
  \min_{\mathcal{Y}}\frac{1}{2}\|\mathcal{B}-\mathcal{Y}\|_{\operatorname{F}}^2, ~\text{s.~t.~} \mathcal{Y}=\operatorname{ifft}(\widehat{\mathcal{Y}},[],3),  
\end{equation*}
where 
$\widehat{\mathcal{Y}}=\widehat{\mathcal{B}}(:,:,1)+\sum_{n=2}^{L}(\widehat{\mathcal{B}}(:,:,n)+\widehat{\mathcal{B}}(:,:,N+2-n))$. $\widehat{\mathcal{B}}=\operatorname{fft}(\mathcal{B},[],3)$, means applying the fast Fourier transform along the sample (3-rd) dimension for the embedding features from different views. 
The obtained low-frequency components incorporate graph similarity into embedding feature learning, resulting in a smooth representation.
\end{definition}
Besides, the term $\sum_{v=1}^{V}\|\tilde{\mathbf{B}}-\mathbf{B}_v\|_{\operatorname{F}}^2$ is included to ensure ISC~\cite{wen2021unified}, where $\tilde{\mathbf{B}}$ is the average embedding feature.

Overall, the framework of S$^2$MVTC is formulated as:
\begin{equation}
\begin{aligned}
&\min_{\{\mathbf{B}_v,\mathbf{U}_v\}_{v=1}^{V},\tilde{\mathbf{B}}}\sum_{v=1}^v(\frac{\lambda}{2}||\mathbf{U}_v||_{\operatorname{F}}^2+\frac{1}{2}||\mathbf{B}_v-\mathbf{U}_v\phi(\mathbf{X}_v)||_{\operatorname{F}}^2 \\
&\qquad\qquad\qquad+\frac{\beta}{2}||\tilde{\mathbf{B}}-\mathbf{B}_v||_{\operatorname{F}}^2) +\tau\|\mathcal{B}\|_{\text{TLFA}} \\
&\text{s.~t. } \tilde{\mathbf{B}}(:,n),\mathbf{B}_v(:,n)\in\mathbb{S}, n=1,\cdots, N, v=1,\cdots,V.
\end{aligned}
\label{eq:alm4}
\end{equation}
Here $\mathcal{B}=\Omega(\mathbf{B}_1,\cdots, \mathbf{B}_V)$, where $\Omega$ represents the operation of stacking each feature and rotating and the inverse operator $\mathbf{B}_v=\Omega_v^{-1}(\mathcal{B})$.

To make the above optimization problem separable, we introduce an auxiliary variable $\mathcal{Y}$, leading to the reformulation as follows:
\begin{equation}
\begin{aligned}
&\min_{\{\mathbf{B}_v,\mathbf{U}_v\}_{v=1}^{V},\tilde{\mathbf{B}}}\sum_{v=1}^v(\frac{\lambda}{2}||\mathbf{U}_v||_{\operatorname{F}}^2+\frac{1}{2}||\mathbf{B}_v-\mathbf{U}_v\phi(\mathbf{X}_v)||_{\operatorname{F}}^2 \\
&+\frac{\beta}{2}||\tilde{\mathbf{B}}-\mathbf{B}_v||_{\operatorname{F}}^2 +\frac{1}{2}||\mathcal{B}-\mathcal{Y}||_{F}^{2}+\tau||\mathcal{Y}||_{\text{TLFA}} \\
&\text{s.~t. } \tilde{\mathbf{B}}(:,n),\mathbf{B}_v(:,n)\in\mathbb{S}, n=1,\cdots, N, 
v=1,\cdots,V. 
\end{aligned}
\label{eq:alm1}
\end{equation}

\subsection{Solutions}
This optimal problem in \cref{eq:alm1} can be solved through an alternating optimization approach, where each parameter is updated individually while keeping the others fixed.\\
\textbf{Update} $\mathbf{U}_v$. Fixing other variables, the subproblem of $\mathbf{U}_v$ can be rewritten as:
\begin{equation}
\min_{\mathbf{U}_v}\frac{1}{2}\|\mathbf{B}_v-\mathbf{U}_v\phi(\mathbf{X}_v)\|_{\operatorname{F}}^2+\frac{\lambda}{2}\|\mathbf{U}_v\|_{\operatorname{F}}^2.
\label{eq:alm2}
\end{equation}
By taking the derivative of the equation and setting it equal to zero, we can obtain:
\begin{equation}
\mathbf{U}_v=(\mathbf{B}_v\phi^{\operatorname{T}}(\mathbf{X}_v))(\phi(\mathbf{X}_v)\phi^{\operatorname{T}}(\mathbf{X}_v)+\lambda \mathbf{I})^{\dagger},
\label{solution:U}
\end{equation}
where $\phi(\mathbf{X}_v)\phi^{\operatorname{T}}(\mathbf{X}_v)$ can be pre-calculated outside the main loop to decrease the computational cost. 
Therefore, the storage and computational complexity are $O(MN)$ and  $O(\max(KMN,M^3))$, respectively.\\
\textbf{Update} $\mathbf{B}_v$.
The subproblem of $\mathbf{B}_v$ can be rephrased as:
\begin{equation}
\begin{aligned}
&\min_{\mathbf{B}_v}\frac{1}{2}\|\mathbf{B}_v-\mathbf{U}_v\phi(x_{v})\|_{F}^{2}+\frac{\beta}{2}\|\tilde{\mathbf{B}}-\mathbf{B}_v\|_{\operatorname{F}}^2 +\frac{1}{2}\|\mathcal{B}-\mathcal{Y}\|_{\operatorname{F}}^2,   \\
& \text{s.t. }  \mathbf{B}_v(:,n)\in\mathbb{S}, n=1,\cdots, N.
\end{aligned}
\label{eq:alm3}
\end{equation}
Similarly, by taking the derivative of the equation and setting it equal to zero, we can obtain
\begin{equation}
\begin{aligned}
\mathbf{B}_v={\text{normalize}}(\beta\tilde{\mathbf{B}}+\Omega_v^{-1}(\mathcal{Y})+{\mathbf{U}_v\phi(\mathbf{x}_v))/(\beta+2)},
\end{aligned}
\label{solution:B}
\end{equation}
where $\Omega_v^{-1}(\mathcal{Y})=\mathbf{Y}_v\in\mathbb{R}^{K\times N}$. $\operatorname{normalize}$ is the $z$-score normalization function, which can be achieved by the Matlab command ``normalize".
In this case,
both the storage and  computational complexity of updating $\mathbf{B}_v$ are $O(KN)$.\\
\textbf{Update} $\mathcal{Y}$. The subproblem of $\mathcal{Y}$ is rewritten as:
\begin{equation}
\begin{aligned}
\min_{\mathcal{Y}} \tau\|\mathcal{Y}\|_{\text{TLFA}}+\frac{1}{2}\|\mathcal{B}-\mathcal{Y}\|_{\operatorname{F}}^2.
\end{aligned}
\label{solution:Y}
\end{equation}
According to the definition of $\|\mathcal{Y}\|_{\text{TLFA}}$, a closed-form solution for $\mathcal{Y}$ can be obtained by choosing the low-frequency component in the Fourier domain. The details are outlined in  \cref{alg:algorithm2}.
The  computational and storage complexity are $O(KVN\log(N))$ and $O(KNV)$, respectively.\\
\textbf{Update} $\tilde{\mathbf{B}}$.
The subproblem of updating $\tilde{\mathbf{B}}$ is:
\begin{equation}
\begin{aligned}
\min_{\tilde{\mathbf{B}}}\sum_{v=1}^{V}\frac{\beta}{2}\|\tilde{\mathbf{B}}-\mathbf{B}_v\|_{\operatorname{F}}^2
\text{s.t. } \tilde{\mathbf{B}}(:,n)\in\mathbb{S}, n=1,\cdots, N.
\end{aligned}
\end{equation}
The solution of $\tilde{\mathbf{B}}$ is:
\begin{equation}
   \tilde{\mathbf{B}}=\operatorname{normalize}(\frac{1}{V}\sum_{v=1}^{V}\mathbf{B}_v),
   \label{solution:bB}
\end{equation}
which needs computational and storage complexity of $O(KN)$ and $O(KNV)$, respectively.

After obtaining the consistent fused embedding feature $\tilde{\mathbf{B}}$.
The final clustering results can be obtained by the following clustering structure learning framework:
\begin{equation}
\begin{aligned}
\min_{\mathbf{D},\mathbf{G}}\|\tilde{\mathbf{B}}-\mathbf{D}\mathbf{G}\|_{\operatorname{F}}^2 \text{~s.~t. } \sum_{c=1}^{C}\mathbf{G}(c,:)=\textbf{1}, \mathbf{G}\in\{0,1\},
\end{aligned}
\label{eq:CG}
\end{equation}
where $\mathbf{D}$ is the clustering center and $\mathbf{G}$ is the clustering indicator matrix. 

Overall, the method S$^2$MVTC is summarized in \cref{alg:algorithm3}. 
According to the parameters analysis in the algorithm, the main storage and computational complexities of S$^2$MVTC are $O(\max(KVN,MN))$ and $O(\max(KMNT,M^3T))$, respectively, where $T$ represents the total number of iterations in the S$^2$MVTC.

In addition, the theoretical convergence can be well guaranteed. Our problem is bounded due to the summation of norms with positive penalty parameters. Furthermore, the exact minimum points of each subproblem can be achieved, implying that each subproblem exhibits a monotone decrease. The proposed algorithm converges according to the convergence theorem in~\cite{rudin1976principles} (theorem 7.29).
\begin{algorithm}[t]
	\caption{Updating $\mathcal{Y}$}	
           	\begin{algorithmic}
			\STATE \textbf{Input:} $\mathbf{B}$,  low frequency parameter $L$.
                \STATE \textbf{Output:}  $\mathcal{Y}$;
                \STATE $\widehat{\mathcal{B}}\leftarrow\mathrm{fft}(\mathcal{B},[],3)$;
                \STATE $\widehat{\mathcal{Y}}(:,:,1)) =\widehat{\mathcal{B}}(:,:,1))$;
           \FOR{$n = 2$ to $L$}
            \STATE $\widehat{\mathcal{Y}}(:,:,n) = \widehat{\mathcal{B}}(:,:,n)$;
              \STATE $\widehat{\mathcal{Y}}(:,:,N+2-n) =\operatorname{conj}(\widehat{\mathcal{Y}}(:,:,n))$.
             \ENDFOR
            \STATE $\mathcal{Y}=\mathrm{ifft}(\widehat{\mathcal{Y}},[],3)$.
	\end{algorithmic}	
\label{alg:algorithm2}
\end{algorithm}
\begin{algorithm}[t]
	\caption{S$^2$MVTC}	
           	\begin{algorithmic}
			\STATE \textbf{Input:} Anchor graphs $\phi(\mathbf{X}_v), v=1,\cdots,V$,  low frequency parameter $L$,  regularization parameters $\beta$, $\lambda$.
            \STATE \textbf{Initialize:}  $\tau=1$, Maximum iterations $T$=7.
                \FOR{$t = 1$ to $T$}
                \FOR{$v = 1$ to $V$}
                    \STATE Update $\mathbf{U}_v$ via \cref{solution:U};
                    \STATE Update $\mathbf{B}_v$ via \cref{solution:B};                
                \ENDFOR
                  \STATE Update $\mathcal{Y}$ via \cref{solution:Y};
                            \STATE Update $\tilde{\mathbf{B}}$ via \cref{solution:bB};
                  \STATE $\tau=\tau*1.1$.
                      \ENDFOR
                \STATE Apply \cref{eq:CG} to find the 
                the clustering indicator matrix $\mathbf{G}$.
                \STATE \textbf{Output:} Clustering result.
         \end{algorithmic}	
\label{alg:algorithm3}
\end{algorithm}

\section{Experiments}
\subsection{Experimental settings}
\begin{figure}[htp!]
\centering
\begin{subfigure}[b]{0.245\textwidth}
\includegraphics[width=\textwidth]{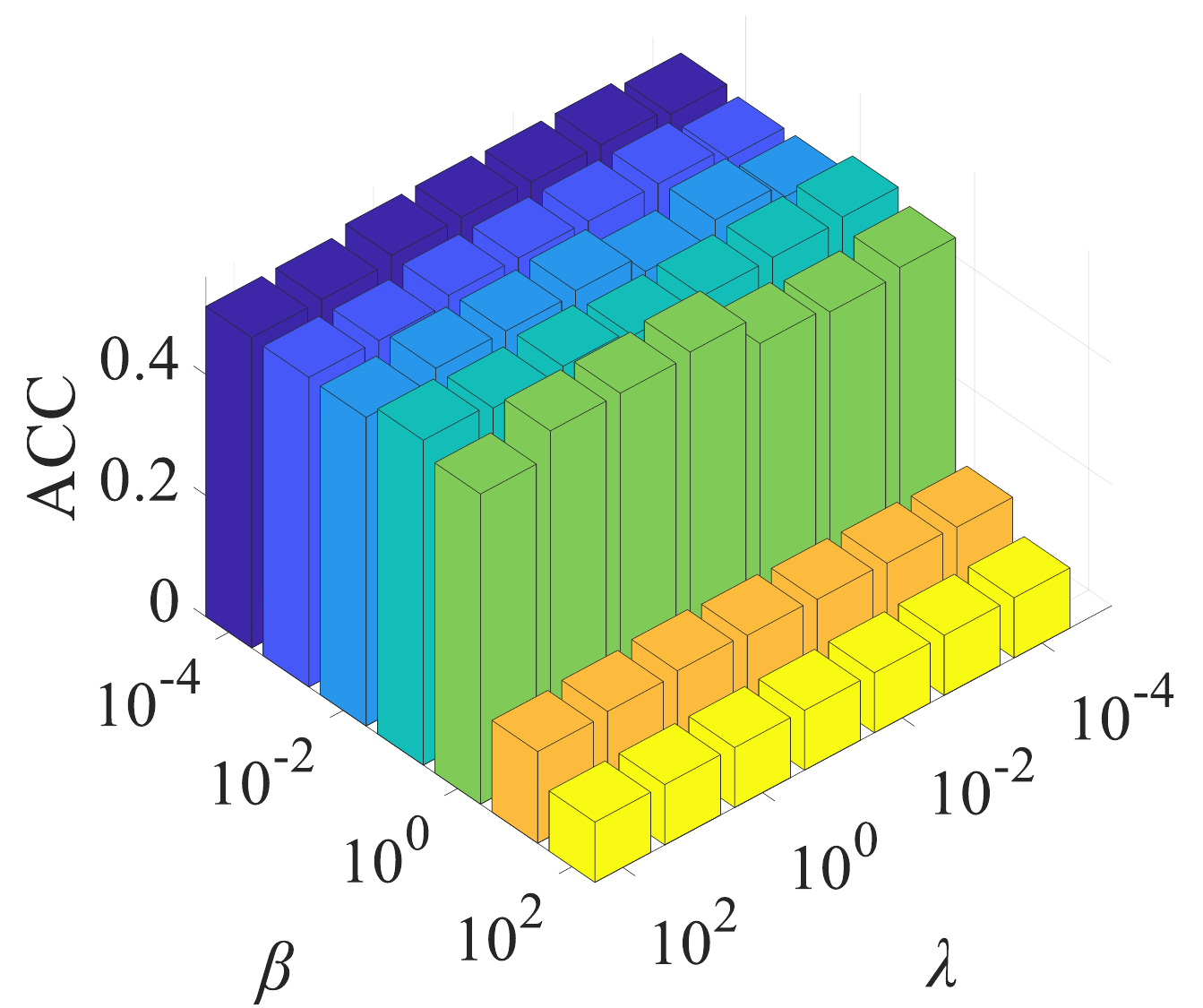}
\caption{$L = 16$, $M = 1000$}
\label{fig:first11}
\end{subfigure}%
\begin{subfigure}[b]{0.245\textwidth}
\includegraphics[width=\textwidth]{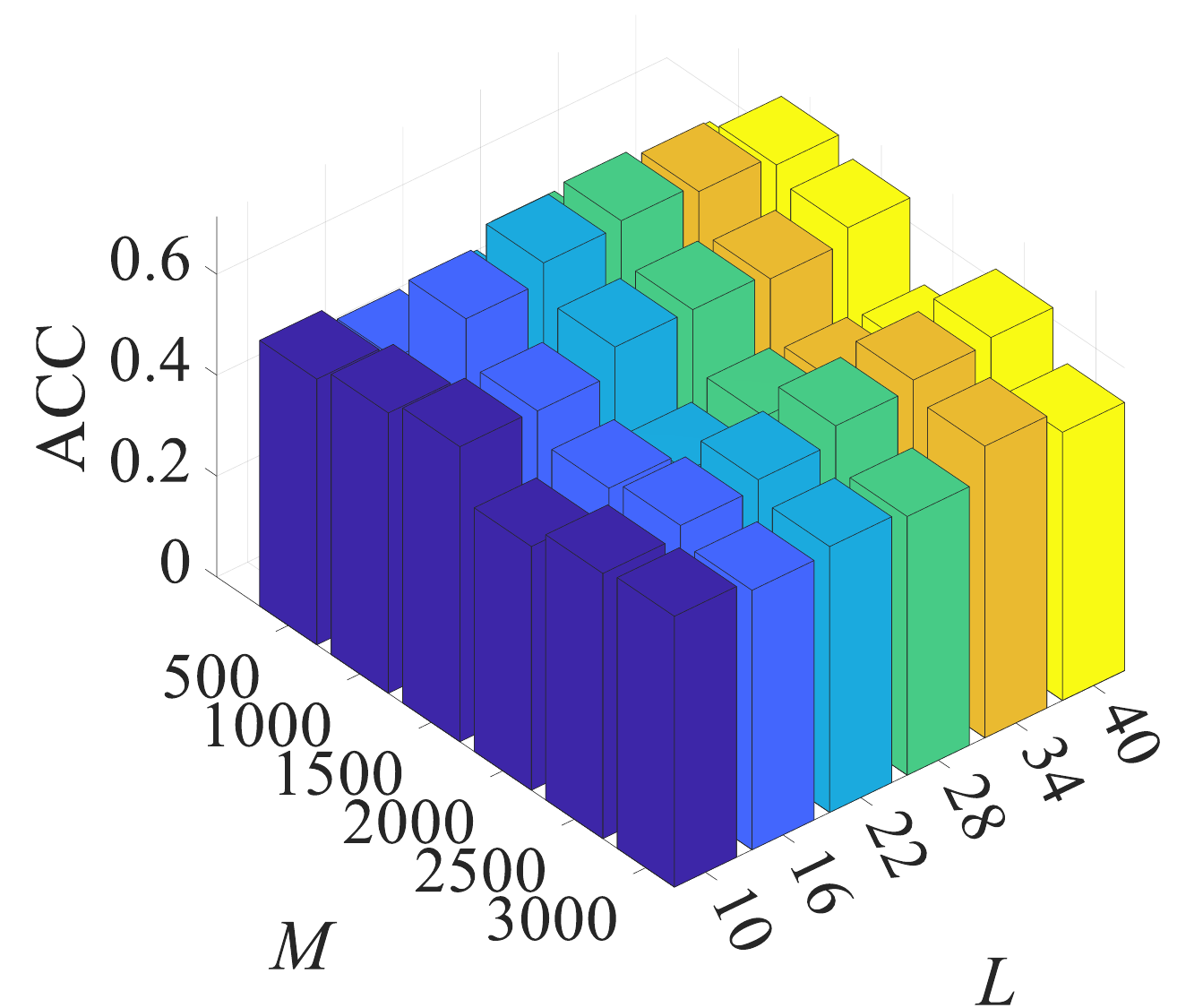}
\caption{$\beta = 1$, $\lambda = 10$}
\label{fig:second22}
\end{subfigure}%
\captionsetup{justification=centering}
\caption{The change of clustering performance as parameters $\lambda$, $\beta$, and $L$, $M$ on Caltech102.}  
\vspace{-0.8cm}
\label{fig:lbml}
\end{figure}

\textbf{Multi-view Datasets}:
Six well-known large-scale multi-view datasets, including \textbf{CCV }~\cite{icmr11:consumervideo}, \textbf{NUS-WIDE-OBJ}~\cite{nus-wide-civr09}, \textbf{Caltech102}~\cite{li_andreeto_ranzato_perona_2022}, \textbf{AwA}~\cite{lampert2014attribute}, \textbf{Cifar-10}~\cite{2012Learning},  \textbf{YoutubeFace\textunderscore sel}~\cite{2011Face} evaluate the effectiveness of S$^2$MVTC, where the statistical information is presented in  \cref{tab:datasets}. \\
\textbf{Compared Clustering Algorithms}.
To compare clustering performance, we select $k$-means and seven state-of-the-art methods: 
binary multi-view clustering (BMVC) [2018, TPMAI]~\cite{zhang2018binary}, 
fast multi-View anchor-correspondence clustering (FMVACC) [2022, NeurIPS]~\cite{wang2022align}, 
one-pass multi-view clustering (OPMC) [2021, ICCV]~\cite{liu2021one}, 
fast multi-view clustering via ensembles (FastMICE) [2023, TKDE]~\cite{huang2023fast}, 
scalable and parameter-free multiview graph clustering (SFMC) [2020, TPMAI]~\cite{li2020multiview}, 
fast self-guided multi-view subspace clustering (FSMSC) [2023, TIP]~\cite{chen2023fast}, 
fast parameter-free multi-view subspace clustering with consensus anchor guidance (FPMVS-CAG) [2021, TIP]~\cite{wang2021fast}.
One low-rank tensor-based multi-view method: 
scalable low-rank MERA based multi-view clustering (sMERA-MVC) [2023, TMM]~\cite{10225718}.
 All tests were conducted on a desktop computer equipped with a 3.79GHz AMD Ryzen 9 3900X CPU and 64GB RAM, using MatLab 2021b.\\
 \begin{table}[t!]
\scalebox{0.66}{
  \centering
  \begin{tabular}{ccccc}
    \toprule
   Datasets  	                   &$\#$Sample &$\#$Cluster&$\#$View  &$\#$ Feature\\
    \midrule
CCV                             & 6773   & 20  & 3 & 20,20,20                      \\
Caltech102                  & 9144   & 102 & 6 & 48,40,254,1984,512,928        \\
NUS-WIDE-OBJ                    & 30000  & 31  & 5 & 65,226,145,74,129         \\
AwA                             & 30475  & 50  & 6  & 2688,2000,252,2000,2000,2000  \\
Cifar-10                        & 50000  & 10  & 3   &   512,2048,1024                 \\
YoutubeFace\textunderscore sel  & 101499 & 31  & 5   & 64,512,64,647,838               \\
    \bottomrule
  \end{tabular}
}
    \caption{Multi-view datasets uesd in our experiments.}  
    \label{tab:datasets}
  \vspace{-0.5cm}

\end{table}
\textbf{Evaluation Metrics}.
To assess the performance in our experiments, eight standard evaluation metrics are utilized, including Accuracy (ACC), Normalized Mutual Information (NMI), Purity, F-score, Precision (PRE), Recall (REC), Adjusted Rand Index (ARI), and CPU Time. These metrics, except for CPU Time, are designed such that higher values signify superior clustering performance. \\
\textbf{Parameter Sensitivity Analysis}.
In our experiment, we employed a grid search strategy to determine the optimal choices for all parameters. S$^2$MVTC has a fixed parameter $M$, which is the size of the anchor, and three free parameters: low frequency parameter, denoted as $L$, and two balance parameters, $\lambda$ and $\beta$. We fixed two of these parameters and adjusted the third through an exhaustive search. For instance in dataset Caltech102, we initially fixed $L = 16$ and $M = 1000$, and adjusted parameters $\beta$ and $\lambda$, where $\beta$ and $\lambda$ took values in \{$10^{-4}$, $10^{-3}$, ..., $10^{2}$\}. The clustering results are shown in  \cref{fig:lbml}\subref{fig:first11}, where the performance remains stable with $\beta$ in the range of $[10^{-4},1]$, with ACC close to 55.47\%. We observed that the best clustering performance can be achieved within a wide range of  $\lambda$, specifically $\lambda \in [10^{-1},10^{1}]$. Additionally, we fixed $\beta = 1$ and $\lambda = 10$, and selected parameter $M$ from \{500, 1000, ..., 3000\} and parameter $L$ from \{10, 16, ..., 40\}. In \cref{fig:lbml}\subref{fig:second22}, performance is stable for $L$ within the $[16,22]$ range, achieving an ACC of approximately 55.47\%. Notably, $M$ was consistently set to 1000 for all test datasets. To ensure more consistent anchor selection in random sampling, the dataset is sorted in ascending order.

\begin{table*}[htbp]
\small

\centering
   \renewcommand{\arraystretch}{0.64}
\begin{tabular}{c|c|c|c|c|c|c|c|c|c}

\toprule \toprule \noalign{\smallskip}

Methods &     ACC &    NMI &    Purity &    F-score &    PRE &     REC &    ARI&     CPU Time (s) &    Speedup \\
    \midrule
\multicolumn{10}{c}{\textbf{CCV} ($\beta$=0.1, $\lambda$=$10^{-5}$, $L$=18)} \\
    \midrule
$k$-means   & 0.1339 & 0.0890 & 0.1680 & 0.0930 & 0.0717 & 0.1321 & 0.0201 & \textbf{1.12} & 1\texttimes    \\
BMVC (2018)      & 0.2263 & 0.2051 & 0.2612 & 0.1390 & 0.1400 & 0.1380 & 0.0870 & 1.95          & 0.57\texttimes \\
SFMC (2020)      & 0.1503 & 0.0904 & 0.1562 & 0.1133 & 0.0637 & 0.5105 & 0.0127 & 17.73         & 0.06\texttimes \\
OPMC (2021)      & 0.1955 & 0.1738 & 0.2309 & 0.1176 & 0.1011 & 0.1407 & 0.0546 &  \underline{1.14}          & 0.98\texttimes \\
FPMVS-CAG (2021) & 0.2284 & 0.1670 & 0.2489 & 0.1333 & 0.1212 & 0.1481 & 0.0749 & 11.62         & 0.10\texttimes \\
FMVACC (2022)    & 0.1919 & 0.1505 & 0.2308 & 0.1143 & 0.1201 & 0.1091 & 0.0632 & 31.70         & 0.03\texttimes \\
FastMICE (2023)  & 0.2076 & 0.1623 & 0.2421 & 0.1244 & 0.1264 & 0.1225 & 0.0720 & 17.99         & 0.06\texttimes \\
FSMSC (2023)     & 0.2167 & 0.1799 & 0.2529 & 0.1273 & 0.1348 & 0.1206 & 0.0773 & 4.50          & 0.25\texttimes \\
sMERA-MVC (2023) &  \underline{0.4554} &  \underline{0.5002} &  \underline{0.5191} &  \underline{0.3406} &  \underline{0.3508} &  \underline{0.3310} &  \underline{0.3015} & 3.81          & 0.29\texttimes \\
S$^2$MVTC &
\textbf{0.5479} &
\textbf{0.6119} &
\textbf{0.5491} &
\textbf{0.4784} &
\textbf{0.3871} &
\textbf{0.6260} &
\textbf{0.4386} &
  1.61 &
  0.70X\\

   \midrule

\multicolumn{10}{c}{\textbf{Caltech102} ($\beta$=1, $\lambda$=10, $L$=16)} \\

    \midrule
$k$-means   & 0.1926 & 0.4083 & 0.3822          & 0.1697 & 0.2390 & 0.2390 & 0.1528 & 40.38          & 1.00\texttimes \\
BMVC (2018)      & 0.3000 & 0.5096 & 0.4850          & 0.2383 & 0.3496 & 0.1807 & 0.2234 & 7.59           & 5.32\texttimes \\
SFMC (2020)      & 0.2440 & 0.3216 & 0.2967          & 0.0547 & 0.0288 & 0.5479 & 0.0012 & 59.55          & 0.68\texttimes \\
OPMC (2021)      & 0.2474 & 0.4848 & 0.4509          & 0.2255 & 0.3456 & 0.1673 & 0.2110 & 31.67          &  \underline{1.28\texttimes} \\
FPMVS-CAG (2021) & 0.3118 & 0.4167 & 0.3786          & 0.2530 & 0.1742 &  \underline{0.4652} & 0.2211 & 162.77         & 0.25\texttimes \\
FMVACC (2022)    & 0.2440 & 0.3216 & 0.2967          & 0.0547 & 0.0288 & 0.5479 & 0.0012 & 57.57          & 0.70\texttimes \\
FastMICE (2023)  & 0.2408 & 0.4897 & 0.4738          & 0.2042 & 0.3606 & 0.1424 & 0.1913 & 280.79         & 0.14\texttimes \\
FSMSC (2023)     & 0.2757 & 0.5118 & 0.4974          & 0.2196 & 0.3951 & 0.1521 & 0.2072 & 16.75          & 2.41\texttimes \\
sMERA-MVC (2023) &  \underline{0.4523} &  \underline{0.8053} & \textbf{0.7128} &  \underline{0.3585} &  \underline{0.5858} & 0.2582 &  \underline{0.3472} & 58.27          & 0.69\texttimes \\
S$^2$MVTC  & \textbf{0.5547} & \textbf{0.8316} &  \underline{0.6667} & \textbf{0.6577} & \textbf{0.6764} & \textbf{0.6400} & \textbf{0.6481} & \textbf{3.56} & 11.34\texttimes\\

      \midrule

\multicolumn{10}{c}{\textbf{NUS-WIDE-OBJ} ($\beta$=1, $\lambda$=$10^{-3}$, $L$=16)} \\

    \midrule
$k$-means   & 0.1227 & 0.1050 & 0.2387 & 0.0849 & 0.1070 & 0.0704 & 0.0390 & 60.04          & 1.00\texttimes \\
BMVC (2018)      & 0.1573 & 0.1351 & 0.2476 & 0.0972 & 0.1138 & 0.0849 & 0.0483 & 28.19          & 2.13\texttimes \\
SFMC (2020)      & 0.1330 & 0.0312 & 0.1356 & 0.1132 & 0.0604 & 0.9066 & 0.0005 & 87.09          & 0.69\texttimes \\
OPMC (2021)      & 0.1438 & 0.1523 & 0.2670 & 0.0973 & 0.1236 & 0.0802 & 0.0524 &  \underline{22.56}          & 2.66\texttimes \\
FPMVS-CAG (2021) & 0.2044 & 0.1338 & 0.2348 & 0.1389 & 0.1124 & 0.1817 & 0.0697 & 75.76          & 0.79\texttimes \\
FMVACC (2022)    & 0.1226 & 0.1113 & 0.2230 & 0.0765 & 0.1048 & 0.0602 & 0.0341 & 84.44          & 0.71\texttimes \\
FastMICE (2023)  & 0.1574 & 0.1540 & 0.2611 & 0.1033 & 0.1380 & 0.0825 & 0.0610 & 44.41          & 1.35\texttimes \\
FSMSC (2023)     & 0.1903 & 0.1326 & 0.2263 & 0.1433 & 0.1164 & 0.1866 & 0.0748 & 30.88          & 1.94\texttimes \\
sMERA-MVC (2023) &  \underline{0.6135}          & \textbf{0.8212} & \textbf{0.8244} &  \underline{0.5996}          & \textbf{0.7230} &  \underline{0.5122}          &  \underline{0.5786}          & 46.15         & 1.30\texttimes \\
S$^2$MVTC      & \textbf{0.6395} &  \underline{0.7440} &  \underline{0.7329}          & \textbf{0.6189} &  \underline{0.6280} & \textbf{0.6101} & \textbf{0.5949} & \textbf{8.35} & 7.19\texttimes\\
      \midrule

\multicolumn{10}{c}{\textbf{AwA} ($\beta$=0.1, $\lambda$=0.03, $L$=9)} \\
  
    \midrule
$k$-means   & 0.0832 & 0.0951 & 0.0984 & 0.0459 & 0.0366 & 0.0615 & 0.0171 & 802.95          & 1.00\texttimes  \\
BMVC (2018)      & 0.1049 & 0.1253 & 0.1120 & 0.0533 & 0.0510 & 0.0558 & 0.0296 &  \underline{32.71}           & 24.55\texttimes \\
SFMC (2020)      & 0.0465 & 0.0288 & 0.0474 & 0.0464 & 0.0238 & 0.9587 & 0.0009 & 129.57          & 6.20\texttimes  \\
OPMC (2021)      & 0.0928 & 0.1193 & 0.1111 & 0.0456 & 0.0449 & 0.0463 & 0.0224 & 182.61          & 4.40\texttimes  \\
FPMVS-CAG (2021) & 0.0949 & 0.1012 & 0.0997 & 0.0603 & 0.0394 & 0.1282 & 0.0255 & 50.81           & 15.80\texttimes \\
FMVACC (2022)    & 0.0769 & 0.0906 & 0.0988 & 0.0381 & 0.0403 & 0.0361 & 0.0164 & 640.66          & 1.25\texttimes  \\

FastMICE (2023)  & 0.0905 & 0.1119 & 0.1114 & 0.0458 & 0.0482 & 0.0435 & 0.0241 & 239.61          & 3.35\texttimes  \\

FSMSC (2023)     & 0.1054 & 0.1197 & 0.1259 & 0.0526 & 0.0563 & 0.0493 & 0.0314 & 42.51           & 18.89\texttimes \\

sMERA-MVC (2023) &  \underline{0.6201} &  \underline{0.7820} & \textbf{0.6888} & \textbf{0.5677} & \textbf{0.5479} &  \underline{0.5889}          & \textbf{0.5570} & 390.59         & 2.06\texttimes  \\
S$^2$MVTC      & \textbf{0.6288} & \textbf{0.8124} &  \underline{0.6334} &  \underline{0.5872} &  \underline{0.4735} & \textbf{0.7726} &  \underline{0.5748} & \textbf{10.35} & 77.58\texttimes\\
   \midrule

\multicolumn{10}{c}{\textbf{Cifar-10} ($\beta$=$10^{-4}$, $\lambda$=$10^{-4}$, $L$=16)} \\

   \midrule
$k$-means   & 0.8935 & 0.7849 & 0.8935 & 0.8013 & 0.7976 & 0.8050 & 0.7791 & 111.04          & 1.00\texttimes \\
BMVC (2018)      & 0.9944 & 0.9846 & 0.9944 & 0.9889 & 0.9889 & 0.9876 & 0.9876 &  \underline{12.09}           & 9.19\texttimes \\
SFMC (2020)      & 0.9873 & 0.9671 & 0.9873 & 0.9750 & 0.9750 & 0.9750 & 0.9722 & 119.74          & 0.93\texttimes \\
OPMC (2021)      & 0.9764 & 0.9420 & 0.9764 & 0.9541 & 0.9540 & 0.9542 & 0.9490 & 21.33           & 5.21\texttimes \\

FPMVS-CAG (2021) & 0.8969 & 0.9423 & 0.8972 & 0.8953 & 0.8284 & 0.9796 & 0.8822 & 35.30           & 3.15\texttimes \\

FMVACC (2022)    & 0.9646 & 0.9627 & 0.9718 & 0.9588 & 0.9489 & 0.9701 & 0.9540 & 134.57          & 0.83\texttimes \\

FastMICE (2023)  & 0.9920 & 0.9778 & 0.9920 & 0.9842 & 0.9842 & 0.9842 & 0.9824 & 94.05           & 1.18\texttimes \\

FSMSC (2023)     & 0.9954 & 0.9701 & 0.9663 & 0.9564 & 0.9406 & 0.9744 & 0.9512 & 56.46           & 1.97\texttimes \\

sMERA-MVC (2023) & \textbf{1.0000} & \textbf{1.0000} & \textbf{1.0000} & \textbf{1.0000} & \textbf{1.0000} & \textbf{1.0000} & \textbf{1.0000} & 214.23         & 0.52\texttimes \\
S$^2$MVTC      &  \underline{0.9994} &  \underline{0.9983} &  \underline{0.9994} &  \underline{0.9989} &  \underline{0.9989} &  \underline{0.9989} &  \underline{0.9988} & \textbf{11.13} & 9.98\texttimes\\

    \midrule

\multicolumn{10}{c}{\textbf{YoutubeFace\_sel} ($\beta$=0.1, $\lambda$=0.005, $L$=19)} \\
  
     \midrule
$k$-means   & 0.1171 & 0.1025 & 0.2723 & 0.0757 & 0.1068 & 0.0586          & 0.0131 & 1167.73         & 1.00\texttimes  \\

BMVC (2018)      & 0.2815 & 0.2818 & 0.3691 & 0.1169 & 0.1911 & 0.0842          & 0.0658 & \textbf{36.12}  & 32.33\texttimes \\
SFMC (2020)      & 0.2843 & 0.0567 & 0.2870 & 0.1665 & 0.0912 & 0.9514          & 0.0039 & 465.44         & 2.51\texttimes  \\
OPMC (2021)      & 0.2574 & 0.2470 & 0.3467 & 0.1139 & 0.1777 & 0.0838          & 0.0600 & 184.88          & 6.32\texttimes  \\
FMVACC (2022)    & 0.2418 & 0.2192 & 0.3329 & 0.0868 & 0.1598 & 0.0596          & 0.0402 & 524.17          & 2.23\texttimes  \\
FPMVS-CAG (2021) & 0.2510 & 0.2440 & 0.3405 & 0.1305 & 0.1538 & 0.1134          & 0.0591 & 483.83          & 2.41\texttimes  \\
FastMICE (2023)  & 0.3029 & 0.2708 & 0.3859 & 0.1189 & 0.2119 & 0.0826          & 0.0723 & 120.32          & 9.71\texttimes  \\
FSMSC (2023)     & 0.2398 & 0.0332 & 0.2689 & 0.1583 & 0.0895 & \textbf{0.6820} & 0.0002 & 174.06          & 6.71\texttimes  \\
sMERA-MVC (2023) &  \underline{0.5381} &  \underline{0.6780} &  \underline{0.7223} &  \underline{0.4279} &  \underline{0.6228} &  \underline{0.3260}          &  \underline{0.3906} & 321.29          & 3.63\texttimes  \\
S$^2$MVTC     & \textbf{0.5732} & \textbf{0.8066} & \textbf{0.7943} & \textbf{0.4308} & \textbf{0.7012} & 0.3109 & \textbf{0.3977} &  \underline{40.09} & 29.13\texttimes \\
\midrule 
    \bottomrule
  \end{tabular}
\caption{The comparison clustering performance on CCV, Caltech102, NUS-WIDE-OBJ, AwA,  Cifar-10 and YoutubeFace\_sel datasets.}  
  \label{tab:results}
\end{table*}

 \begin{figure*}[h!]
 \centering
\includegraphics[width=1\textwidth]{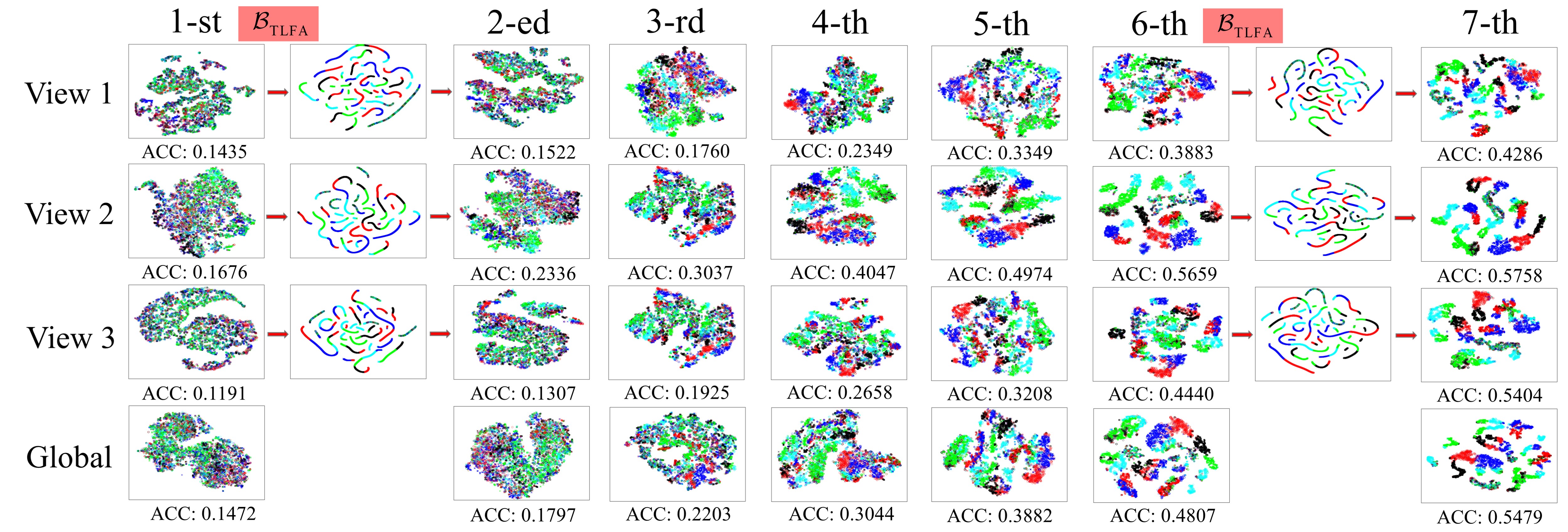}
\caption{ The embedding feature learning process on the CCV dataset. 
 From left to right, the nines columns represent the embedding features after the 1-st iteration, the 1-st TLFA operator, the 2-ed to 6-th iteration, the 6-th TLFA operator, and the 7-th iteration, respectively. 
Each feature is visualized using t-SNE~\cite{van2008visualizing}: View 1 for SIFT features, View 2 for STIP features, View 3 for MFCC features, and ‘Global' for the fusion of these features.}
\label{fig:process}
\vspace{-0.5cm}
 \end{figure*}

\subsection{Clustering performance analysis}
 \cref{tab:results} reports the clustering performance of various methods on six large-scale multi-view datasets, where the best and second-best are highlighted in \textbf{bold} and \underline{underlined}, respectively.  
From the table, it is evident that sMERA-MVC and S$^2$MVTC significantly outperform others across various metrics such as ACC, NMI, Purity, F-score, PRE, REC, and ARI. 
sMERA-MVC, leveraging advanced tensor networks, excels in exploring both inter-view and intra-view correlations of anchor graphs among views simultaneously. This allows sMERA-MVC to utilize the most compact structure for identifying global correlations among anchor graphs, which could be a key factor contributing to its superior performance.

In comparison with sMERA-MVC, S$^2$MVTC consistently outperforms it in most scenarios. Notably, S$^2$MVTC demonstrates a significant improvement in clustering performance on the CCV dataset. For instance, the clustering performance of S$^2$MVTC has shown remarkable enhancements of 9.25\%, 11.17\%, and 13.71\% in terms of ACC, NMI, and ARI, respectively.
In the case of the Caltech102, S$^2$MVTC exhibits a performance advantage, achieving a 10.24\% higher ACC compared to sMERA-MVC. For the NUS-WIDE-OBJ and AwA datasets, the performances of both methods are similar. Regarding the YoutubeFace\_sel dataset, which is larger than other large-scale datasets, sFSR-IMVC demonstrates the best performance compared to sMERA-MVC. 

Significantly, in terms of CPU time, , S$^2$MVTC outperforms sMERA-MVC across all large datasets, especially with larger sizes. For instance, it achieves an approximation speedup of 39 times on AwA and 20 times on Cifar-10. Furthermore, BMVC, relying on binary feature representation for clustering, exhibits a CPU time performance at a comparable scale to S$^2$MVTC on large datasets. However, the clustering performance of S$^2$MVTC surpasses BMVC, with ACC improvements of 32.16\% on CCV, 25.47\% on Caltech102, 48.22\% on NUS-WIDE-OBJ, 52.04\% on Cifar-10, and 29.18\% on YoutubeFace\_sel, respectively.

In summary, based on the analysis of these metrics, it can be concluded that the S$^2$MVTC algorithm excels not only in accuracy for multi-view clustering tasks but also significantly outperforms traditional algorithms in terms of computational efficiency, making it suitable for fast large-scale multi-view clustering tasks.

\subsection{Model discussions}
\textbf{Model Analysis}. 
\cref{fig:process} illustrates the embedding feature learning process of S$^2$MVTC on the CCV dataset. 
Compared the 1-st and 3-rd columns, it is evident that after the TLFA operation, the dispersion of sample points increases. This is because that TLFA aims at identifying the compact subspace representation of embedding features across views, known as a smooth representation. During the iterative process, the complementary information captured in this compact subspace further propagates across views. Simultaneously, the exploration of complementary information facilitates the discovery of improved and more smooth subspace representations, ultimately resulting in enhanced clustering performance. For instance, after the 6-th TLFA operation, the clustering performance reaches 54.79\%.\\


\textbf{Ablation Study}.
The S$^2$MVTC model consists primarily of three components: 1. Exploration of ISC; 2. Exploration of IGS; 3. Adoption of a nonlinear anchor graph. To further investigate why S$^2$MVTC performs well, we systematically removed each component while keeping the other two fixed, naming these methods as w/o ISC, w/o IGS, and linear anchor graphs. The results are presented in  \cref{tab:Ablation}.
From the table, it can be observed that removing the exploration of ISC leads to a slight decline in clustering performance, while the removal of the exploration of IGS results in a significant drop in clustering performance. This indicates that the comprehensive exploration of graph similarity among intra-view features effectively, thereby improving clustering performance. 
Furthermore, when the anchor graph is transformed into a linear mapping between anchors and samples, clustering performance decreases with an increase in the number of samples, especially in Cifar-10 and YoutubeFace\_sel.  This is because, with a fixed number of anchors, a nonlinear anchor graph is better at capturing relationships between samples when the sample size is exceptionally large.

 \begin{table}[h]
   \vspace{-0.2cm}
\scalebox{0.75}{
  \centering
 \begin{tabular}{c|c|c|c|c}
 \toprule
\multicolumn{1}{l}{} &  \multicolumn{1}{c}{w/o ISC} &  \multicolumn{1}{c}{w/o IGS} &  \multicolumn{1}{c}{linear anchor graphs} &  \multicolumn{1}{c}{S$^2$MVTC} \\
      \midrule
Dataset          & ACC      & ACC     & ACC      & ACC      \\
    \midrule
CCV              & 54.42 & 15.49 & 47.31 & 54.79 \\
Caltech102       & 51.31 & 17.27 & 44.71 & 55.47 \\
NUS-WIDE-OBJ     & 63.06 & 17.17 & 63.21 & 63.95 \\
AwA              & 51.61 &  8.08 & 58.97 & 62.53 \\
Cifar-10         & 99.94 & 65.92 & 88.25 & 99.94 \\
YoutubeFace\_sel & 53.97 & 11.38 & 40.20 & 57.33 \\
    \bottomrule 
\end{tabular}
}
  \vspace{-0.2cm}
  \caption{Ablation Studies on six large-scale datasets.}
  \label{tab:Ablation}
  \vspace{-0.6cm}
\end{table}

\section{Conclusion}
In this paper, we present a simple yet efficient scalable multi-view tensor clustering, where the intra-view and inter-view correlations are directly learned from the embedding features.  
During the iterative process, the introduced TLFA operator optimizes the embedding features to a compact subspace. This optimization aids in exploring complementary information and its propagation across multiple independent views, ensuring maximum consistency between different views.
Additionally, we incorporate inter-view semantic consistency and a nonlinear anchor graph to enhance clustering performance. 
Numerical experiments on six large-scale multi-view datasets demonstrate that our method outperforms all state-of-the-art methods, showcasing a notable improvement with increasing data size.

{
    \small
    \bibliographystyle{ieeenat_fullname}
    \bibliography{ref}
}


\end{document}